\documentclass[a4paper,conference]{IEEEtran}
\IEEEoverridecommandlockouts

\ifCLASSINFOpdf
  \usepackage[pdftex]{graphicx}
  \DeclareGraphicsExtensions{.pdf,.jpeg,.png}
\else
\fi

\usepackage{cite}
\usepackage{comment}
\usepackage{amsmath,amssymb,amsfonts}
\usepackage{algorithmic}
\usepackage{subcaption}
\usepackage{graphicx}
\usepackage{textcomp}
\usepackage{booktabs}
\usepackage{multirow}
\usepackage[left=1.29cm, right=1.29cm, top=1.9cm, bottom=4.29cm]{geometry}

\usepackage{optidef}
\usepackage{xcolor}
\usepackage{hyperref}
\usepackage{longtable}
\usepackage{gensymb}
\usepackage[]{algorithm2e}

\usepackage{glossaries}
\usepackage{soul}
\usepackage{inconsolata}

 \makeglossaries
\usepackage[font=small]{caption}
\usepackage{enumitem}
\usepackage{anyfontsize}
\usepackage{balance}
\usepackage{fancyhdr}
\makeatletter
\renewcommand{\fnum@figure}{Figure \thefigure}
\makeatother

\title{A Neural ODE Approach to Aircraft Flight Dynamics Modelling}

\vspace{0.5cm}

\newcommand{\specialcell}[2][c]{%
  \begin{small}%
  \begin{tabular}[#1]{@{}c@{}}%
 #2%
  \end{tabular}%
  \end{small}}

\author{%
Gabriel Jarry\IEEEauthorrefmark{1},
Ramon Dalmau\IEEEauthorrefmark{1},
Xavier Olive\IEEEauthorrefmark{2},
Philippe Very\IEEEauthorrefmark{1}

\\[1em]
\begin{tabular}{cc}%
\specialcell{\IEEEauthorrefmark{1}\small EUROCONTROL \\ Aviation Sustainability Unit (ASU) \\
Brétigny-sur-Orge, France} &
  \specialcell{\IEEEauthorrefmark{2}\small ONERA -- DTIS\\Universit\'e de Toulouse\\Toulouse, France} 
\end{tabular}
}

\IEEEaftertitletext{\vspace{-1\baselineskip}}

\begin{document}
\maketitle

\begin{abstract}
Accurate aircraft trajectory prediction is critical for air traffic management, airline operations, and environmental assessment. This paper introduces NODE-FDM, a Neural Ordinary Differential Equations-based Flight Dynamics Model trained on Quick Access Recorder (QAR) data. By combining analytical kinematic relations with data-driven components, NODE-FDM achieves a more accurate reproduction of recorded trajectories than state-of-the-art models such as a BADA-based trajectory generation methodology (BADA4 performance model combined with trajectory control routines), particularly in the descent phase of the flight. The analysis demonstrates marked improvements across altitude, speed, and mass dynamics. Despite current limitations, including limited physical constraints and the limited availability of QAR data, the results demonstrate the potential of physics-informed neural ordinary differential equations as a high-fidelity, data-driven approach to aircraft performance modelling. Future work will extend the framework to incorporate a full modelling of the lateral dynamics of the aircraft.
\end{abstract}

\begin{IEEEkeywords}
Aircraft performance modelling, Neural-ODE, trajectory prediction, physics-informed learning
\end{IEEEkeywords}

\section{Introduction}

Accurate aircraft trajectory prediction is essential to air traffic management (ATM), airline operations, and aviation environmental assessment. Precise trajectory models are indispensable for conflict detection and resolution, sector capacity planning, and trajectory-based operations (TBO). Beyond operational safety and efficiency, high-fidelity predictions enable robust estimates of fuel consumption, emissions, and noise, thereby supporting the development of sustainable aviation strategies.

The most widely adopted reference for aircraft performance modelling is the Base of Aircraft Data (BADA)~\cite{nuic2010user, nuic2010bada}, developed by EUROCONTROL. BADA provides a standardized set of equations and performance parameters to represent aircraft dynamics under nominal conditions. Its analytical formulation offers transparency and robustness, which explains its longstanding use in both operational and research contexts. Nevertheless, such models inevitably rely on simplifying assumptions and manufacturer-provided data, which can lead to divergences between simulated trajectories and those actually flown, particularly in speed profiles and mass dynamics as captured by Automatic Dependent Surveillance-Broadcast (ADS-B) measurements. Alternative open-source frameworks, such as OpenAP~\cite{sun2020openap}, have emerged to complement BADA by providing flexible, data-driven tools for trajectory modelling and validation.

This study introduces NODE-FDM, a novel data-driven Flight Dynamics Model that integrates large-scale Quick Access Recorder (QAR) data with physics-informed neural architectures. We focus first on the vertical and speed dimensions of the trajectory, including altitude profiles, flight path angle, and true airspeed. This paper does not address lateral trajectory management (e.g., route choice and turn dynamics), which we leave for future investigations. We apply the model to the Airbus A320-214 aircraft.

The proposed model combines analytical kinematic relations, regression-based propulsion and aircraft angle modelling, and a Neural Ordinary Differential Equations (ODE) formulation to predict the temporal evolution of altitude and speed with high fidelity. We evaluate its predictive performance against a BADA-based trajectory generation methodology (BADA4 performance model combined with trajectory control routines, hereafter referred to as BADA), focusing on trajectory reconstruction accuracy and suitability for environmental impact assessments. These assessments include developing CO$_2$ and non-CO$_2$ emissions inventories, which often require robust trajectory predictions even when only partial datasets or pre-flight planning information are available.

The remainder of this paper is organized as follows. Section~\ref{sec:sota_trajectory} reviews the literature on aircraft trajectory prediction, covering model-based, data-driven, and hybrid physics-informed approaches, and discusses operational use-cases and open challenges. Section~\ref{sec:data} describes the QAR datasets and preprocessing steps used to train the model. Section~\ref{sec:methodo} details the NODE-FDM methodology, including its modular architecture, state representation, neural and analytical layers, and training objectives. Section~\ref{sec:results} presents quantitative results, benchmark comparisons with BADA, and analysis of trajectory reconstruction and fuel consumption estimation. Section~\ref{sec:discussion} discusses the limitations of the proposed approach, and Section~\ref{sec:conclusion} concludes with a summary of findings and directions for future work.

\section{Literature review}
\label{sec:sota_trajectory}

Trajectory prediction and generation underpin a wide range of ATM and airline operations, from conflict detection and sequencing to environmental assessment and strategic planning. Two complementary paradigms dominate: model-based approaches grounded in aircraft performance and point-mass dynamics, and data-driven methods that learn spatio-temporal regularities from large-scale surveillance and weather datasets. 



Physics-based point-mass formulations and performance databases remain the historical backbone of operational trajectory prediction. The EUROCONTROL BADA family provides aircraft-type parameters and procedures for 4D trajectory computation under forecast conditions~\cite{nuic2010user,nuic2010bada}, while OpenAP extends these capabilities to the research community with an open-source framework~\cite{sun2020openap,sun2022openap}.  Beyond canonical models, the literature formalizes trajectory representation, compression, and optimization using splines, principal component analysis, and connections to optimal control and wavefront propagation under constraints~\cite{delahaye2014mathematical}. Recent efforts by Poll and Schumann refine performance characteristics and cruise fuel modelling from aerodynamic theory and empirical data~\cite{poll2021estimation,poll2021estimation2}. These approaches provide interpretability and controllability, but rely on accurate state and intent estimates and may under-represent operational variability.

Data-driven methods have achieved notable gains in 4D trajectory prediction using supervised learning with surveillance and meteorological features. Early work by Ayhan and Samet demonstrated machine-learning-based prediction of arrival times and trajectories using historical and weather data~\cite{ayhan2016aircraft}. Sequence models then became prevalent: Liu and Hansen proposed a seq2seq long short-term memory (LSTM)~\cite{liu2018predicting}, Pang et al. combined convolutional encoders of convective weather “cubes” with recurrent decoders~\cite{pang2019recurrent}, and Ma and Tian introduced a convolutional neural networks (CNN)–LSTM on ADS-B data~\cite{ma2020hybrid}. Online frameworks fuse point-mass physics, BADA parameters, and ADS-B conformance to update intent and improve estimated time of arrival precision~\cite{zhang2018online}. Generative models address multi-modality and error accumulation: generative adversarial network (GAN)-based forecasters produce full-sequence 4D trajectories~\cite{wu2022long}, variational autoencoders (VAE)/GAN generators improve realism in sparse regimes~\cite{chen2021trajvae}, privacy-preserving LSTM–GANs synthesize flows while mitigating re-identification risk~\cite{rao2020lstm}, and imitation-learning methods such as TrajGAIL learn realistic route choices and flow-level statistics~\cite{choi2021trajgail}. Data-driven models excel at capturing variability and airline-specific practices, though generalization off-distribution and physical guarantees remain challenging~\cite{jarry2024generalization}.

Hybrid strategies combine the adaptability of data-driven approaches with the consistency of physics-based models, either by calibrating physics-based predictors using data or by embedding known dynamics into neural architectures. Calibration work has improved climb predictions by inferring aircraft-specific parameters from radar data~\cite{alligier2018learning}. Physics-informed learning enforces equations of motion and aerodynamic limits via loss functions or architectural constraints. Neural ODEs formalize these approaches as continuous-time dynamical systems, trained efficiently with adjoint methods and evaluated using ODE solvers~\cite{chen2018neural,haber2017stable}. Aerospace applications include learning residual dynamics under icing~\cite{ma2024development}, hypersonic glide vehicle prediction~\cite{lu2024enhanced}, and differentiable predictive control for unmanned aerial vehicles~\cite{park2025data}. Practical challenges remain regarding stiffness, training cost, stability, and hybrid dynamics with discrete modes~\cite{ruthotto2020deep}.

Beyond prediction, ATM applications also require the generation of realistic trajectories and flows with quantified realism and flyability. Statistical density models, such as Gaussian mixture models and vine copulas, combined with dimensionality reduction, reproduce flow statistics and go-around patterns while balancing realism and tractability~\cite{krauth2021synthetic,krauth2023deep}. Evaluation frameworks assess operational realism, statistical coherence, similarity to observations, and flyability via simulator replay~\cite{olive2021framework}. Interactive FPCA-based pipelines support clustering, deformation, and consistent generation for what-if analyses, for instance in noise footprint studies~\cite{jarry2022interactive}. Aviation-focused GANs synthesize approach paths and identify atypical trajectories for safety management~\cite{jarry2019use}, while recent adversarial maximization–minimization methods assemble trajectories from mined manoeuvres under performance constraints, improving fidelity, diversity, and controller consistency~\cite{gui2024novel}.

Overall, model-based methods provide transparency and physical consistency but struggle with latent intent and operational diversity, data-driven predictors capture variability and multi-modality yet must address generalization, uncertainty, and physical plausibility, and physics-guided hybrids, including Neural ODE frameworks, offer a promising middle ground for robustness and realism. Within this spectrum, our approach falls into the physics-guided hybrid category, combining large-scale QAR data with analytical kinematics, regression-based propulsion modeling, and a Neural ODE formulation. Compared with purely model-based approaches such as BADA, it emphasizes accurate reconstruction of altitude and speed trajectories while remaining suitable for environmental impact assessments.

\section{Data sources}
\label{sec:data}

The dataset used in this study is derived from Quick Access Recorder (QAR) data collected on nine different Airbus A320-214 aircraft operated by the airline Aigle Azur over two years. QAR systems provide high-frequency recordings, with the sampling rate depending on the parameter, typically ranging from 1 to 4 Hz. They capture a wide range of flight parameters, including aircraft state variables (e.g., altitude, true airspeed, Mach number), control inputs (e.g., thrust setting, flap configuration, landing gear status), and environmental conditions (e.g., temperature, wind components, runway elevation). Compared with publicly available surveillance data such as ADS-B, or radar data managed by air navigation service providers, QAR recordings offer higher temporal resolution and additional parameters critical for detailed aircraft performance modelling. However, QAR datasets are owned and maintained by airlines and are not routinely shared publicly, which limits access and restricts the ability to generalise models across operators.

The raw QAR data underwent several preprocessing steps to ensure consistency and suitability for machine learning applications. First, data were subsampled to 0.25 Hz (one point every four seconds). Second, all parameters were converted into standard SI units, and statistical measures (mean and standard deviation) were computed across the training set to provide normalisation capabilities for the neural networks. Each flight was then divided into non-overlapping sequences of fixed length (60 time steps, equivalent to four minutes) to form the samples used for training and validation. Table~\ref{tab:qar_features} summarises the features extracted from the QAR data and used in the model.

\begin{table*}[!h]
\centering
\caption{List of parameters used in the model with original QAR units and converted SI units.}
\label{tab:qar_features}
\begin{tabular}{lllll}
\textbf{Feature} & \textbf{Symbol} & \textbf{QAR Unit} & \textbf{SI Unit} & \textbf{Description} \\
\hline
\\
\multicolumn{5}{l}{\textbf{State variables $x(t)$}} \\ \\
altitude & $h$ & ft & m & Standard altitude above mean sea level \\
distance along track & $d$ & nm & m & Along-track distance from departure \\
flight path angle & $\gamma$ & deg & rad & Angle between velocity vector and horizon \\
true airspeed & $V_{\mathsf{TAS}}$ & kt & m/s & Speed relative to the surrounding air mass \\
mass & $m$ & kg & kg & Estimated aircraft mass \\ \\
\hline
\\
\multicolumn{5}{l}{\textbf{Control variables $u(t)$}} \\ \\
selected altitude & $h_{\mathsf{sel}}$ & ft & m & Altitude target set in the FMS/autopilot \\
selected speed & $V_{\mathsf{sel}}$ & kt & m/s & Speed target set in the FMS/autopilot \\
selected vertical speed & $V_{z,\mathsf{sel}}$ & ft/min & m/s & Vertical speed target set in the FMS/autopilot (if set else 0.0) \\
flap setting & $f$ & index & index & Current flap configuration 0,1,2,3,4 (FULL) \\
landing gear status & $g$ & binary & binary & Gear up (0) / down (1) \\
speed brake command & $SB$ & binary & binary & Speed brake retracted (0) / deployed (1) \\ \\
\hline
\\
\multicolumn{5}{l}{\textbf{Context and intermediate variables $e(t)$}} \\ \\
\multicolumn{5}{l}{\textbf{Context variables $e_0(t)$}} \\
outside air temperature & $T_{\mathsf{OAT}}$ & $^\circ$C & K & Ambient static air temperature \\
headwind component & $V_{\parallel}$ & kt & m/s & Wind component along the trajectory \\
crosswind component & $V_{\perp}$ & kt & m/s & Wind component orthogonal to the trajectory \\ \\
\multicolumn{5}{l}{\textbf{Trajectory variables $e_1(t)$}} \\
Mach number & $M$ & -- & -- & Ratio of $V_{\mathsf{TAS}}$ to speed of sound \\
calibrated airspeed & $V_{\mathsf{CAS}}$ & kt & m/s & Airspeed derived from dynamic pressure \\
vertical speed & $V_z$ & ft/min & m/s & Rate of climb or descent \\
ground speed & $V_{\mathsf{GS}}$ & kt & m/s & Speed relative to the ground \\ 
selected altitude difference & $h_{\mathsf{sel}} - h$ & ft & m & Difference between selected and current altitude \\
selected speed difference & $V_{\mathsf{sel}} - V_{\mathsf{CAS}}$ & kt & m/s & Difference between selected speed and CAS \\ \\
\multicolumn{5}{l}{\textbf{Aircraft angles $e_2(t)$}} \\
angle of attack & $\alpha$ & deg & rad & Angle between chord line and relative wind \\
pitch angle & $\theta$ & deg & rad & Angle between aircraft longitudinal axis and horizon \\ \\
\multicolumn{5}{l}{\textbf{Engine parameters $e_3(t)$}} \\
N1 (fan speed) & $\mathsf{N}_1$ & \%RPM & \%RPM & Low-pressure compressor rotation speed \\
fuel flow & $m_{\mathsf{fuel}}$ & kg/h & kg/s & Total fuel consumption rate \\
\hline
\end{tabular}
\end{table*}

The final dataset comprises 5,000 training flights, 500 validation flights, and 500 test flights, distributed across five airframes for training, two for validation, and two for testing, with no overlap of aircraft between the subsets. Each flight is divided into 4 minutes non-overlapping sequences and is represented by a multi-source tensor that groups three categories of variables:
\begin{itemize}
    \item \textbf{state variables $x(t)$} describe the instantaneous state of the aircraft and its motion. They include altitude, along-track distance, true airspeed, flight path angle, and estimated aircraft mass.
    \item \textbf{control variables $u(t)$} reflect the pilot’s or autopilot’s inputs and mode selections. They include speed and altitude selector differences, flap setting, and landing gear status. When the speed target is set in Mach, the QAR also records the equivalent calibrated airspeed target ($V_{sel}$); in this study, only the $V_{sel}$ representation is retained. Similarly, when the aircraft is commanded to follow a constant slope, the QAR does not log the corresponding vertical speed target explicitly; we compute the selected vertical speed ($V_{z,\mathsf{sel}}$) from the recorded ground speed and target slope.
    \item \textbf{context and intermediate variables $e(t)$} provide environmental conditions and derived parameters relevant to aircraft dynamics. They include static air temperature, horizontal wind components, engine parameters (e.g., N1 fan speed, thrust setting), and aerodynamic angles such as angle of attack.
\end{itemize}

\section{Methodology}
\label{sec:methodo}

\begin{figure*}[!ht]
    \centering
    \includegraphics[width=0.9\textwidth]{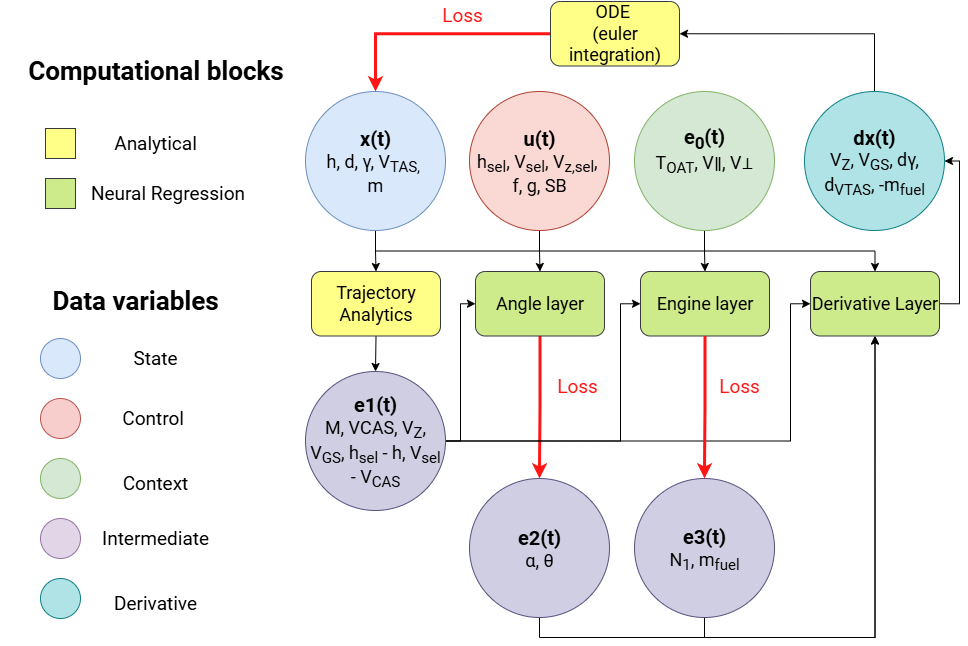}
    \caption{Architecture of the proposed model}
    \label{fig:model}
\end{figure*}

The proposed NODE-FDM is organised as a modular, physics-informed architecture designed to estimate the temporal evolution of aircraft states, as illustrated in Figure~\ref{fig:model}. The model combines analytical equations, regression-based neural layers, and a Neural ODE formulation to ensure both physical consistency and flexibility, with a focus on the vertical profile and speed management. In this  study, the scope is limited to the longitudinal dimensions of the trajectory: altitude, true airspeed, flight path angle, aircraft mass, and along-track distance. These variables play a central role in determining fuel consumption, emissions, and noise, making them particularly relevant for environmental assessment. Lateral dynamics (e.g. track angle and horizontal position) are not included at this stage and are left for future investigation.

The state vector is defined as:
\begin{equation}
x(t) = \{h(t), d(t), \gamma(t), V_{\mathsf{TAS}}(t), m(t)\},
\end{equation}

\noindent  where $h$ denotes the altitude, $d$ the along-track distance, $\gamma$ the flight path angle, $V_{\mathsf{TAS}}$ the true airspeed, and $m$ the aircraft mass. These variables are either logged directly by the QAR from onboard instruments or derived from sensor-based computations available in the aircraft systems. This state vector corresponds to the standard point-mass formulation, widely applied in aircraft trajectory prediction and optimization.

In the implementation, a structured layer is defined and serves as a reusable building block within the model architecture. Each structured layer consists of three main components: an input normalizer, a backbone, and multiple output heads. The inputs are first normalized according to their empirical statistics and concatenated into a fixed-size representation. This representation is processed by the backbone, a two-layer fully connected network with 24 hidden neurons per layer, followed by ReLU activations. The extracted features are then passed to task-specific output heads, each implemented as a single-layer perceptron. This design allows the model to share a common latent representation while adapting flexibly to heterogeneous output variables (continuous or binary). Finally, the outputs of each head are denormalized to recover values on the original physical scale. We deliberately excluded dropout layers from our Neural ODE architecture. Unlike in discrete architectures (e.g., ResNets), applying standard dropout within continuous-time models is known to destabilize training, as the stochastic masking conflicts with the smooth dynamics required for stable ODE integration \cite{leeneural}.

The complete model is composed of four main layers:

\begin{itemize}
    \item a \textbf{trajectory layer} (analytical): deterministic computation of vertical speed, Mach number, and ground speed from current state and difference of selected vs real speed and altitude.
    
    \item an \textbf{angle layer} (neural network): a structured layer predicting angle-related parameters such as angle of attack and pitch from control inputs and aircraft states.
    
    \item an \textbf{engine layer} (neural network): a structured Layer predicting engine-related parameters such as fuel flow and fan speed ($\mathsf{N}_1$) from control inputs and aircraft states.

    \item a \textbf{derivative layer} (neural network): a structured layer that estimates the derivatives $\mathrm{d}V_{\mathsf{TAS}}, \; \mathrm{d}\gamma$ of the state vector, while the remaining derivatives are obtained through direct usage of existing variables: $V_{\mathsf{GS}}$, $-m_{\mathsf{fuel}}$, and $V_z$. Based on the concatenated state, control, context and intermediate variables, this block provides $\mathrm{d}V_{\mathsf{TAS}}$ and $\mathrm{d}\gamma$, which are then integrated with an ODE solver (Euler scheme in this work) to generate continuous-time trajectories.

\end{itemize}

The \textbf{trajectory layer} provides deterministic conversions from the state variables and context conditions to additional kinematic variables, ensuring consistency with physical laws. Specifically:
\begin{align}
V_z &= V_{\mathsf{TAS}} \cdot \sin(\gamma), \\
M   &= \frac{V_{\mathsf{TAS}}}{a}, \qquad 
    a = \sqrt{\gamma_{air} \, R \, T_{\mathsf{OAT}}}, \\
V_{\mathsf{GS}} &= V_{\mathsf{TAS}} - V_{\parallel},
\end{align}
where $V_z$ is the vertical speed, $M$ the Mach number, $a$ the local speed of sound, $\gamma_{air}$ the ratio of specific heats, $R$ the perfect gas constant, $T_{\mathsf{OAT}}$ the outside air temperature, and $V_{\parallel}$ the headwind component.

The evolution of the state vector is modelled as:
\begin{equation}
\frac{dx}{dt} = f_{\theta}(x(t), u(t), e(t)),
\end{equation}

\noindent where $x(t)$ denotes the aircraft state vector at time $t$, $u(t)$ represents the control variables, and $e(t)$ corresponds to the context and intermediate variables. $f_{\theta}$ is represented by the \texttt{derivative} layer (a neural network). We compute the predicted trajectories through numerical integration of the ODE using an explicit Euler solver, implemented via the \texttt{odeint} function from the \texttt{torchdiffeq} library \cite{chen2018neural}.

The model is trained by minimizing the difference between predicted and observed trajectories using a composite loss function. 
This loss combines errors across several groups of outputs, namely the state variables, aircraft angles, and engine parameters:

\begin{equation}
\mathcal{L} = \sum_{i \in \mathcal{C}} \alpha_i \,\ell_i(y_i^{\text{pred}}, y_i^{\text{true}}),
\end{equation}

\noindent where $\ell_i$ is the error metric for feature $i$, $\alpha_i$ is a weighting coefficient, and $\mathcal{C} = \{x(t), \; e_2 (t), \; e_3 (t)\}$ (see Table~\ref{tab:qar_features}). In other words, the loss function directly compares the model’s predicted variables with those recorded by the QAR. The resulting error signals are then used to adjust the weights of the neural ODE, gradually improving its prediction accuracy. 

In this paper, $\ell_i$ corresponds to a mean squared error (MSE) and 
the weighting coefficients $\alpha_i$ are chosen inversely proportional to the empirical standard deviation of each variable, in order to mitigate differences in scale and absolute value. This formulation allows balancing between accuracy in integrated state values, and regressed engine and aircraft variables.

The model is trained on one GPU (RTX A6000 Ada Gen) using \texttt{PyTorch} for 1000 epochs with batch size of 500 and checkpoints on the validation set. Optimization was performed with the \texttt{AdamW} optimizer (\(\texttt{lr}=1\times10^{-4}, \ \texttt{weight\_decay}=1\times10^{-4}\) corresponding to L2 regularization).

\section{Results}
\label{sec:results}

\begin{table*}[ht]

\centering
\caption{Comparison of NODE-FDM and BADA performance across phases. Metrics are reported as mean (standard deviation), 
based on 849,490 trajectory points across 500 flights.}
\label{tab:metric}
\begin{tabular}{llccc|ccc}
\toprule
\multirow{2}{*}{parameter} & \multirow{2}{*}{phase} & \multicolumn{3}{c}{NODE-FDM} & \multicolumn{3}{c}{BADA} \\
\cmidrule(lr){3-5} \cmidrule(lr){6-8}
 & & MAE & MAPE (\%) & ME & MAE & MAPE (\%) & ME \\
\midrule
altitude [m] & All phases & 61.90 (145.54) & 2.49 (333.62) & 0.47 (158.15) & 167.47 (402.75) & 5.03 (315.60) & -75.45 (429.61) \\
 & Climb & 93.03 (121.98) & 3.36 (569.52) & 57.64 (142.16) & 214.69 (180.19) & 6.14 (269.86) & 175.79 (218.31) \\
& Level Flight & 8.70 (33.50) & 0.26 (6.16) & -1.51 (34.57) & 12.32 (67.80) & 0.28 (9.20) & -8.26 (68.42) \\
& Descent & 182.04 (246.23) & 7.95 (467.70) & -49.49 (302.19) & 560.15 (713.73) & 17.39 (637.75) & -509.53 (750.69) \\

\midrule

true air speed [m/s] & All phases & 1.35 (2.03) & 0.75 (1.27) & -0.04 (2.44) & 2.73 (6.16) & 1.64 (5.18) & -1.34 (6.61) \\
 & Climb & 1.70 (1.93) & 0.95 (1.27) & 0.87 (2.42) & 3.31 (3.33) & 2.02 (2.66) & 1.86 (4.31) \\
 & Level Flight & 0.70 (0.75) & 0.32 (0.43) & -0.04 (1.03) & 0.90 (1.62) & 0.45 (1.41) & -0.62 (1.75) \\
 & Descent & 2.82 (3.33) & 1.76 (2.02) & -0.92 (4.27) & 7.32 (11.55) & 4.64 (10.15) & -6.51 (12.02) \\
 
\midrule
$\gamma$ [deg] & All phases & 0.24 (0.42) & N/A & 0.05 (0.48) & 0.46 (1.01) & N/A & 0.05 (1.11) \\
 & Climb & 0.42 (0.51) & N/A & 0.04 (0.65) & 0.64 (1.11) & N/A & 0.03 (1.28) \\
 & Level Flight & 0.08 (0.16) & N/A & 0.03 (0.18) & 0.07 (0.29) & N/A & 0.03 (0.30) \\
 & Descent & 0.51 (0.60) & N/A & 0.09 (0.78) & 1.36 (1.49) & N/A & 0.09 (2.01) \\
 
\midrule

mass [kg]   & All phases & 64.89 (70.44) & 0.10 (0.11) & 7.26 (95.50) & 154.89 (122.70) & 0.25 (0.20) & 152.90 (125.17) \\
 & Climb & 22.98 (31.34) & 0.04 (0.05) & 7.12 (38.21) & 60.22 (58.35) & 0.09 (0.09) & 59.65 (58.94) \\
 & Level Flight & 74.91 (73.32) & 0.12 (0.12) & 6.06 (104.65) & 170.43 (119.16) & 0.27 (0.20) & 168.63 (121.69) \\
 & Descent & 77.29 (74.27) & 0.13 (0.12) & 10.80 (106.64) & 202.96 (130.73) & 0.34 (0.22) & 199.05 (136.61) \\
\bottomrule
\end{tabular}
\end{table*}

\begin{table*}[ht]
\centering
\caption{Comparison of consumption prediction performance (Predicted vs. BADA). Metrics are reported as mean (standard deviation), 
based on 849,490 trajectory points across 500 flights.}
\label{tab:consumption}
\begin{tabular}{lccc|ccc}
\toprule
\multirow{2}{*}{parameter} & \multicolumn{3}{c}{NODE-FDM} & \multicolumn{3}{c}{BADA} \\
\cmidrule(lr){2-4} \cmidrule(lr){5-7}
 & MAE [kg] & MAPE (\%) & ME [kg] & MAE [kg] & MAPE (\%) & ME [kg] \\
\midrule
consumption &  83.95 (79.65) & 1.54 (1.13) & 3.01 (115.74) & 163.52 (129.68) & 3.03 (1.75) & -149.67 (145.47) \\
\bottomrule
\end{tabular}
\end{table*}
To generate physically consistent benchmark trajectories from operational flight data, we used a BADA-based methodology combining the \texttt{pyBADA 4.2} library with the \texttt{TCL.py} (Trajectory Calculation Library) module, together with control inputs mapped from QAR data. In this setup, the BADA4 aircraft performance model was instantiated for the A320-214, while trajectory propagation relied on the TCL routines \verb+accDec_time+, \texttt{constantSpeedLevel}, \verb+constantSpeedRating_time+, and \verb+constantSpeedROCD_time+, applied over fixed 4-second intervals. For each QAR (Quick Access Recorder) record, the current aircraft state (altitude, true airspeed, Mach number, temperature deviation from ISA, configuration, and mass) was extracted and mapped to the corresponding BADA4 input variables. Calibrated airspeed and Mach number were derived from standard atmosphere relationships, while along-track wind and speed-brake deployment were also incorporated to ensure consistency with the recorded operational conditions. Importantly, the evaluation presented here concerns the entire BADA-based trajectory generation pipeline, not only the underlying BADA4 performance model.

Depending on the identified flight phase (climb, cruise, or descent) and the difference between commanded and actual speeds, the appropriate TCL routine was selected: \verb+constantSpeedLevel+ for cruise, \verb+constantSpeedRating_time+ for climb or descent at constant speed, \verb+accDec_time+ when speed adjustments were required, and \verb+constantSpeedROCD_time+ when a vertical speed target had been engaged. At each integration step, the most recent simulated state, i.e., altitude, airspeed, rate of climb or descent, and mass, was fed back into the next computation, thereby ensuring a continuous and dynamically consistent simulation of the trajectory.

For testing purposes, complete trajectories (from take-off to landing) were generated in simulation by propagating the aircraft state from the first recorded point $p_{0}$ after take-off until the last point before landing. The propagation was performed over the same time horizon as the corresponding QAR trajectory, using the full sequence of recorded control inputs (speed targets, vertical mode selections, and configuration changes). This approach ensures that the simulated trajectory reproduces the temporal structure of the operational flight while remaining entirely model-driven, thereby providing a consistent benchmark for evaluating prediction performance.

The comparative analysis of NODE-FDM and BADA-based simulations, presented in Table~\ref{tab:metric}, highlights several clear patterns. Overall, NODE-FDM achieves lower mean absolute errors across altitude, airspeed, and mass compared with BADA, with the most significant gains observed during the climb and descent phases. For instance, the altitude error of NODE-FDM remains below 100~m in climb and around 200~m in descent, whereas BADA shows noticeably larger deviations, with errors of about 200~m in climb and more than 550~m in descent.

Similarly, NODE-FDM provides a closer reproduction of the recorded airspeed dynamics, with typical deviations around 1.35~m/s, while BADA shows larger discrepancies with more than 2.70~m/s mean absolute error. 

\begin{figure}[htbp]
    \centering
    \includegraphics[width=0.95\linewidth]{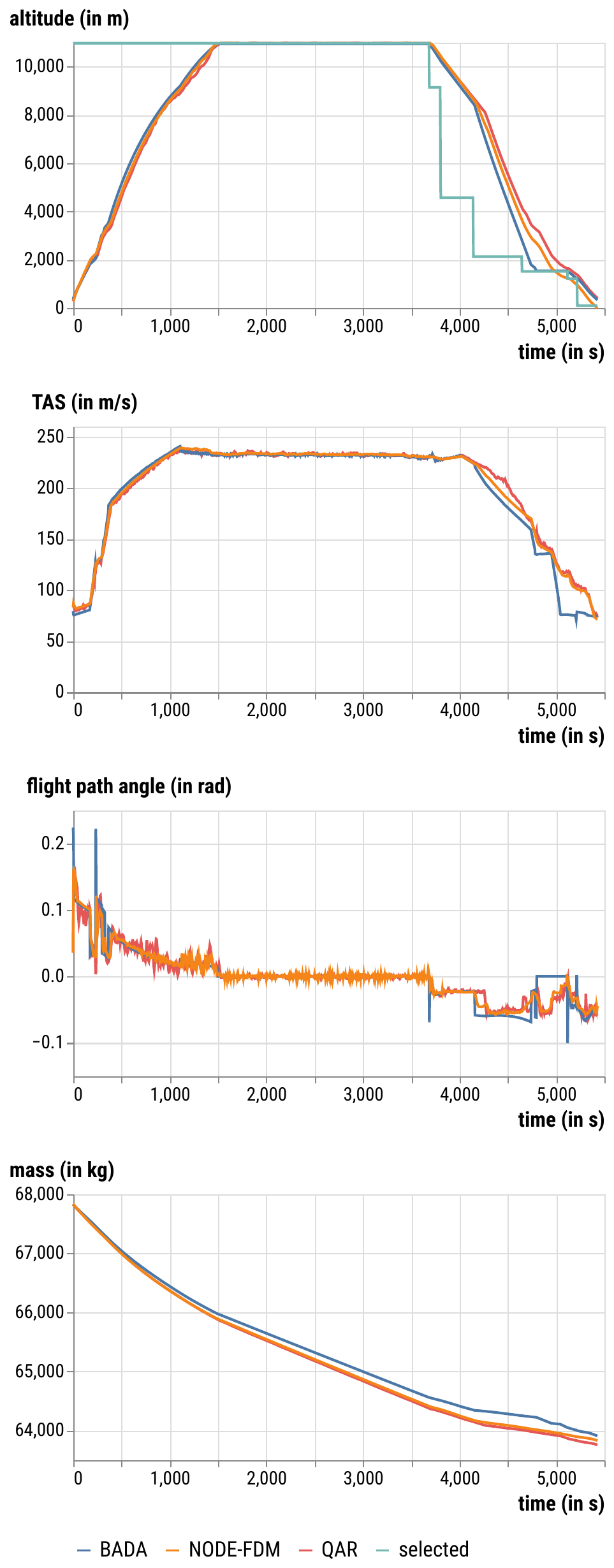}
    \caption{Example of trajectory reproduction from QAR data. 
    The figure compares the recorded QAR trajectory (red) with the NODE-FDM simulation (orange) 
    and the BADA-based benchmark (blue) across altitude, true airspeed, flight path angle, 
    and mass. The selected altitude is shown as a blue-green line. 
    Time is represented on the x-axis in seconds (one point every 4~s).}
    \label{fig:traj_example}
\end{figure}

When examining the flight path angle, NODE-FDM exhibits a smaller dispersion of errors, although both models encounter challenges in reproducing this variable accurately, reflecting the inherent sensitivity of $\gamma$ to minor variations in altitude and airspeed. Regarding mass evolution, NODE-FDM closely follows the recorded values, with errors on the order of sixty kilograms, whereas BADA’s estimates remain consistently biased by one hundred and fifty kilograms. This difference should be interpreted with caution, as BADA relies on performance coefficients calibrated on aircraft in factory new conditions rather than the specific QAR dataset employed here.

It is important to emphasise that BADA is a well-established and physically consistent performance model, widely adopted in operational and simulation contexts. Its higher errors in this comparison do not indicate a lack of validity; rather, they reflect methodological differences. Here, we are evaluating the entire BADA-based trajectory generation pipeline, which couples the BADA4 performance model with trajectory control routines using operational inputs, rather than the performance model in isolation. BADA is not intended to replicate individual recorded flights with exact control input fidelity, but to provide a reliable and generalisable representation of aircraft performance across a broad range of conditions. NODE-FDM, by contrast, is trained directly on the same QAR data used for evaluation, which naturally gives it an advantage in reproducing the observed dynamics.

Taken together, these results demonstrate that NODE-FDM can achieve high-fidelity trajectory reproduction when provided with detailed flight data, while BADA remains a robust and reliable reference model, particularly well suited for large-scale studies and operational applications where generality, consistency, and transparency are essential.

The general trends observed in the quantitative analysis are further illustrated by the example trajectory shown in Figure~\ref{fig:traj_example}. In this case, NODE-FDM more closely reproduces the recorded QAR dynamics across altitude, speed, and mass, whereas BADA exhibits larger deviations, particularly during descent. Nonetheless, both models capture the overall trajectory evolution and phase transitions, confirming the consistency between the aggregated statistical results and individual trajectory examples. The comparison presented in Table~\ref{tab:consumption} illustrates the implications of different modelling approaches for estimating fuel consumption in the context of environmental impact assessments.

NODE-FDM provides a markedly more precise estimate of absolute consumption, with lower mean absolute and percentage errors compared to BADA. This improvement arises from the model’s integration of aircraft- and trajectory-specific dynamics, as well as a direct loss on the fuel flow, which reduces biases. In particular, NODE-FDM accounts for the degradation of engine performance due to in-service fleet ageing, which BADA does not capture. Consequently, BADA tends to systematically under-estimate absolute fuel burn \cite{jarry2025aircraftfuelflowmodelling}.

Nevertheless, despite this limitation, BADA successfully reproduces relative consumption trends across flights. The correlation between BADA-derived consumption and the actual reference remains strong, demonstrating its utility when the objective is to capture variability and patterns at a large scale, such as for trend analysis, relative comparisons, or traffic scenario studies.

\section{Discussion and Limitations }
\label{sec:discussion}

A first limitation of the present work is that NODE-FDM is restricted to the vertical–longitudinal dynamics of the trajectory, i.e., the evolution of altitude and speed along the aircraft’s longitudinal axis, expressed through true airspeed. In this formulation, the aircraft is effectively represented in a body-fixed reference frame aligned with its nose, so only motion in the vertical plane is captured. Consequently, lateral behaviour, including variations in track angle, banking, and turn dynamics, is not represented. The framework therefore cannot yet support fully four-dimensional trajectory prediction, nor operational applications in which lateral manoeuvres and route choices play a decisive role. Extending the model to incorporate lateral dynamics represents a natural and necessary avenue for future research.

A second limitation is that the proposed approach is not strictly constrained by physical laws beyond the analytical kinematic relations embedded in the trajectory layer. While the Neural ODE formulation produces smooth and coherent trajectories, it does not guarantee adherence to aerodynamic or performance bounds. Unrealistic behaviours may therefore arise under unusual control inputs or when the model is applied outside the distribution of the training dataset. Introducing stronger physics-based constraints or regularisation terms could mitigate these risks and improve generalisation.

Finally, reliance on QAR data represents a structural limitation. QAR datasets provide rich, high-resolution information essential for model development and evaluation, but they are not widely accessible at scale. In the present work, the approach has been applied to a single aircraft type and model instance rather than across a full fleet, limiting the scope of the conclusions. This restricted availability poses challenges for developing models intended to generalise across fleets, operators, or aircraft types. Broader access to standardised, large-scale operational datasets would greatly facilitate such transferability.

\section{Conclusion}
\label{sec:conclusion}

This paper has introduced NODE-FDM, a Neural ODE-based Flight Dynamics Model, illustrated using high-resolution Quick Access Recorder (QAR) data from Airbus A320-214 aircraft. By combining analytical kinematic relations with data-driven components, the model reproduces recorded trajectories more closely than the widely used Base of Aircraft Data (BADA) benchmark, particularly in altitude, speed, and mass evolution. Comparative analysis shows that NODE-FDM substantially reduces mean errors during climb and descent phases while maintaining coherence in level flight, highlighting its potential as a high-fidelity trajectory reconstruction tool.

Beyond improvements in accuracy, these results demonstrate the benefits of hybrid modelling strategies that embed physical structure within learning architectures. Nevertheless, several limitations remain: the absence of lateral dynamics, the lack of strict physical constraints on the learned components, and the restricted availability of QAR data, which limits large-scale applicability across fleets. Addressing these challenges represents a promising avenue for future research, particularly by extending the framework to fully four-dimensional trajectories, integrating stronger physics-based constraints, and leveraging more accessible datasets. Future work will specifically investigate the use of automatic dependent surveillance-broadcast (ADS-B) data, coupled with a simplified version of NODE-FDM, to address scalability and enable deployment at fleet or network level while balancing fidelity and generalisability.

\balance
\bibliographystyle{IEEEtran}
\bibliography{references}

\end{document}